\documentclass[sigconf]{acmart}

\usepackage{booktabs} % For formal tables
\usepackage{balance}
\usepackage{url,graphicx}
\usepackage{mathrsfs, amsfonts}
\usepackage{multirow}

\usepackage{enumitem}
\setlist[itemize]{noitemsep, topsep=0pt}

\usepackage{balance}
\usepackage{graphicx}
\usepackage{caption}
\usepackage{subcaption}

\usepackage{algorithm}
\usepackage{algorithmic}

\usepackage[flushleft]{threeparttable}

\begin{document}
\title[Neural Rating Regression with Abstractive Tips Generation...]{Neural Rating Regression with Abstractive Tips Generation for Recommendation}
\titlenote{The work described in this paper is substantially supported by grants 
		from the Research Grant Council of the Hong Kong Special Administrative 
		Region, China (Project Code: 14203414) and the Microsoft Research Asia Urban Informatics Grant FY14-RES-Sponsor-057.
		This work is also affiliated with the CUHK MoE-Microsoft Key Laboratory 
		of Human-centric Computing and Interface Technologies.}

%\author{
%	Piji Li$^{\dag}$ \ \ \ \  Zihao Wang$^{\dag}$  \ \ \ \ Zhaochun Ren$^{\ddag}$ \ \ \ \  Lidong Bing$^{\S}$ \ \ \  \ Wai Lam$^{\dag}$ \\
%	$^{\dag}$Department of Systems Engineering and Engineering Management,\\
%	The Chinese University of Hong Kong, Hong Kong\\
%	$^{\ddag}$Data Science Lab, JD.com, Beijing, China\\
%	$^{\S}$ AI Lab, Tencent Inc., Shenzhen, China\\
%	$^{\dag}$\{pjli, zhwang, wlam\}@se.cuhk.edu.hk, $^{\ddag}$renzhaochun@jd.com, $^{\S}$lyndonbing@tencent.com
%}

\author{Piji Li}
\affiliation{%
	\institution{Department of Systems Engineering and Engineering Management, \\
		The Chinese University of Hong Kong.}
	%\city{Hong Kong} 
	%\postcode{999077}
}
\email{pjli@se.cuhk.edu.hk}

\author{Zihao Wang}
\affiliation{%
	\institution{Department of Systems Engineering and Engineering Management, \\
		The Chinese University of Hong Kong.}
	%\city{Hong Kong} 
	%\postcode{999077}
}
\email{zhwang@se.cuhk.edu.hk}

\author{Zhaochun Ren}
\affiliation{%
	\institution{Data Science Lab, JD.com,}
	\streetaddress{North Star Century Center,}
	\city{Beijing} 
	\country{China}}
\email{renzhaochun@jd.com}

\author{Lidong Bing}
\affiliation{
	\institution{AI Lab, Tencent Inc.}
	\streetaddress{Hi-tech Park, Nanshan District,}
	\city{Shenzhen} 
	\country{China}}
\email{lyndonbing@tencent.com}

\author{Wai Lam}
\affiliation{%
	\institution{Department of Systems Engineering and Engineering Management, \\
		The Chinese University of Hong Kong.}
	%\city{Hong Kong} 
	%\postcode{999077}
}
\email{wlam@se.cuhk.edu.hk}

\renewcommand{\shortauthors}{Piji Li et al.}

\fancyhead{}
\settopmatter{printacmref=false, printfolios=false}

\copyrightyear{2017} 
\acmYear{2017} 
\setcopyright{acmlicensed}
\acmConference{SIGIR '17}{August 07-11, 2017}{Shinjuku, Tokyo, Japan}\acmPrice{15.00}\acmDOI{10.1145/3077136.3080822}
\acmISBN{978-1-4503-5022-8/17/08}

\begin{abstract}
Recently, some E-commerce sites launch a new interaction box called Tips on their mobile apps. Users can express their experience and feelings or provide suggestions using short texts typically several words or one sentence.
In essence, writing some tips and giving a numerical rating are two facets of a user's product assessment action, expressing the user experience and feelings.
Jointly modeling these two facets is helpful for designing a better recommendation system.
While some existing models integrate text information such as item specifications or user reviews into user and item latent factors for improving the rating prediction, no existing works consider tips for improving recommendation quality.
We propose a deep learning based framework named \textbf{NRT} which can simultaneously predict precise ratings and generate abstractive tips with good linguistic quality simulating user experience and feelings.
For abstractive tips generation, gated recurrent neural networks are employed to ``translate'' user and item latent representations into a concise sentence.
%All the neural parameters and the latent factors are learnt by a multi-task learning approach in an end-to-end training paradigm.
Extensive experiments on benchmark datasets from different domains show that NRT achieves significant improvements over the state-of-the-art methods.
Moreover, the generated tips can vividly predict the user experience and feelings.
\end{abstract}



% We no longer use \terms command


%\settopmatter{printacmref=false}

\maketitle

%\section{Introduction}

\section{Introduction}
\label{section1}

\begin{figure}[!t]
	\centering
	\includegraphics[width=1\columnwidth]{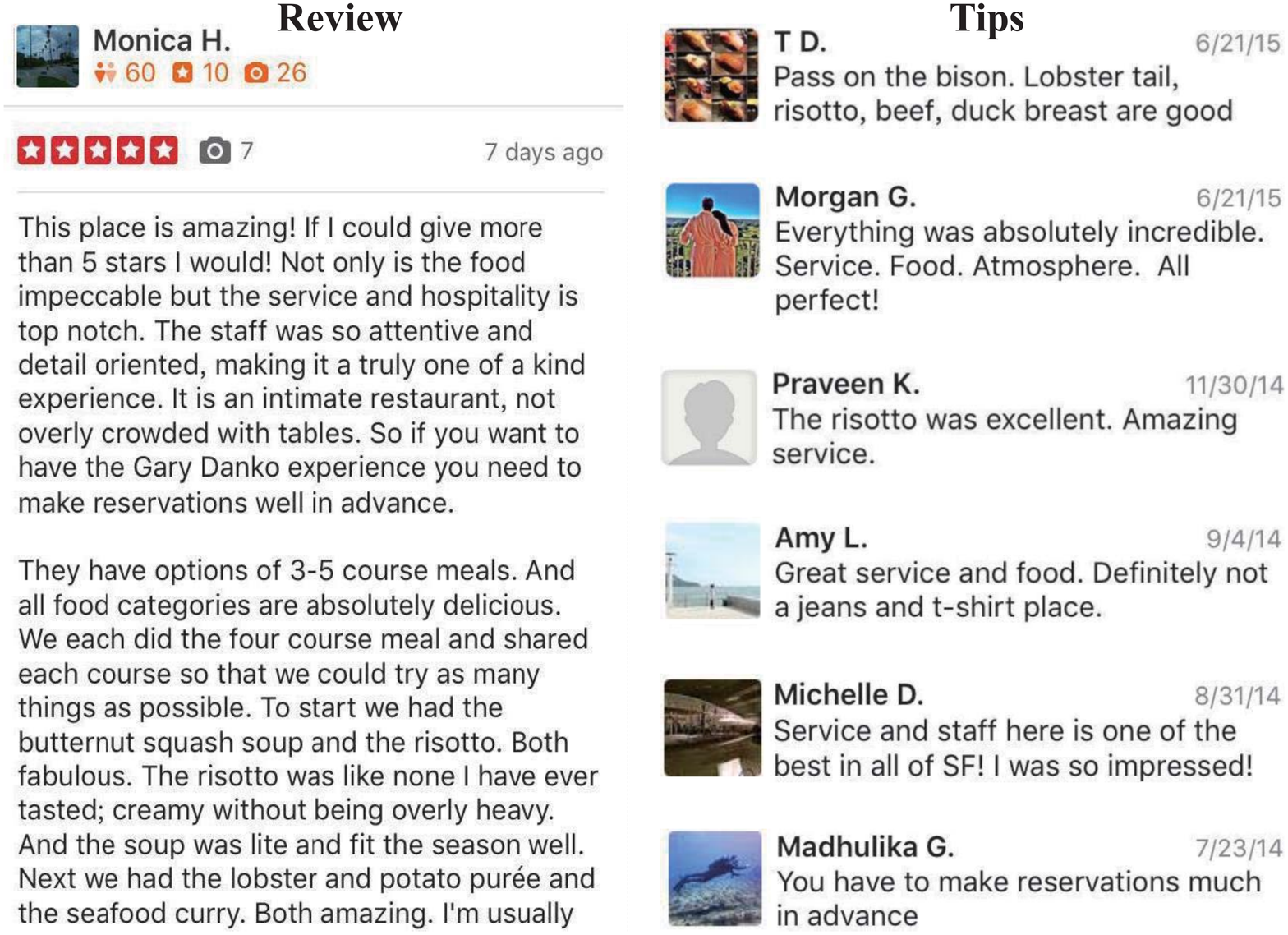}
	\caption{
		Examples of reviews and tips selected from the restaurant ``Gary Danko'' on Yelp. Tips are more concise than reviews and can reveal user experience, feelings, and suggestions with only a few words. Users will get conclusions about this restaurant immediately after scanning the tips with their mobile phones.
	}
	\label{fig:front}
	\vspace{0mm}
\end{figure}

With the explosive growth of Internet information, recommendation systems have been playing an increasingly important role in on-line E-commerce and applications in a variety of areas, including music streaming service such as  Spotify\footnote{http://www.spotify.com} and Apple Music, movie rating such as IMDB\footnote{http://www.imdb.com}, video streaming service such as Netflix and Youtube, job recommendation such as LinkedIn\footnote{http://www.linkedin.com}, and product recommendation such as Amazon.
Many recommendation methods are based on Collaborative Filtering (CF) which mainly makes use of historical ratings \cite{sarwar2001item,mnih2007probabilistic,koren2009matrix,koren2008factorization,shi2010list,lee2001algorithms,marlin2003modeling}.
Recently, some approaches also consider text information in addition to the rating data \cite{wang2011collaborative,mcauley2013hidden,ling2014ratings,almahairi2015learning,zheng2017joint,rensocial2017}.
After some investigations, we observe that the text information in most recommendation tasks can be generally classified into two types: item specifications \cite{wang2011collaborative,wang2015collaborative,wang2016collaborative} and user reviews \cite{mcauley2013hidden,xu2014collaborative,ling2014ratings,xu2015unified,almahairi2015learning,zheng2017joint,rensocial2017}. Item specifications are the text information for describing the attributes or properties of the items.
For example, in article recommendation such as CiteULike\footnote{http://www.citeulike.org}, it refers to titles and abstracts of papers. In product recommendation such as Amazon, it refers to product descriptions and technical specification information.
The second type is user reviews which are written by users to explain why they like or dislike an item based on their usage experiences.  Multi-faceted information can be extracted from reviews and used as user preferences or item features, which otherwise cannot be obtained from the overall ratings \cite{chen2015recommender}.
Although both types of text data are found to be useful for the recommendation task, they have some inherent limitations.
Concretely, the former cannot reflect users' experience and preference, and the latter is usually too long and suffers from noise.

Recently, some E-commerce sites such as Yelp\footnote{http://www.yelp.com} launch a new interaction box called \textbf{Tips} on their mobile platforms.
As shown in Figure~\ref{fig:front}, the left column is a review from the user ``Monica H.'', and tips from several other users are shown on the right column.
In the review text, Monica first generally introduced the restaurant, and then narrated her dining experience in detail.
In the tips text, users expressed their experience and feelings plainly using short texts, such as ``The risotto was excellent. Amazing service.''. They also provide some suggestions to other people directly in several words, such as ``You have to make reservations much in advance.''
In contrast to item specifications and user reviews, tips have several characteristics: (1) tips are typically single-topic nuggets of information, and shorter than reviews with a length of about 10 words on average; (2) tips can express user experience, feelings, and suggestions directly; (3) tips can give other people quick insights, saving the time of reading long reviews.
In essence, writing some tips and giving a numerical rating are two facets of a user's product assessment action, expressing the user experience and feelings.
Jointly modeling these two facets is helpful for designing a better recommendation system.

Existing models only integrate text information such as item specifications \cite{wang2011collaborative,wang2015collaborative,wang2016collaborative} and user reviews \cite{mcauley2013hidden,xu2014collaborative,ling2014ratings,xu2015unified,almahairi2015learning,zheng2017joint,rensocial2017} to enhance the performance of latent factor modeling and rating prediction.
To our best knowledge, we are the first to consider tips for improving the recommendation quality.
We aim at developing a model that is capable of  conducting the latent factor modeling and rating prediction, and more importantly, it can generate tips based on the learnt latent factors.
We do not just extract some existing sentences and regard them as tips.
Conversely, we investigate the task of automatically construing a concise sentence as tips, such capability can be treated as simulating how users write tips in order to express their experience and feelings, just as if
they have bought and consumed the item.
Therefore, we named this task \textit{abstractive tips generation}, where ``abstractive'' is a terminology 
from the research of text summarization \cite{lidong15absmds}.

Generating abstractive tips only based on user latent factors and item latent factors is a challenging task.
Recently, gated recurrent neural networks such as Long Short-Term Memory (LSTM) \cite{hochreiter1997long} and Gated Recurrent Unit (GRU) \cite{cho2014learning} demonstrate high capability in text generation related tasks \cite{bahdanau2014neural,rush2015neural}.
Moreover, inspired by \cite{he2017neural,wang2015collaborative}, neural network based models can help learn more effective latent factors when conducting rating prediction and improve the performance of collaborative filtering.
We employ deep learning techniques for latent factor modeling, rating prediction, and abstractive tips generation.
For abstractive tips generation, gated recurrent neural networks are employed to ``translate'' a user latent factor and an item latent factor into a concise sentence to express user experience and feelings.
For neural rating regression, a multilayer perceptron network \cite{rosenblatt1961principles} is employed to project user latent factors and item latent factors into ratings.
All the neural parameters in the gated recurrent neural networks and the multilayer perceptron network as well as the latent factors for users and items are learnt by a multi-task learning approach in an end-to-end training paradigm.

The main contributions of our framework are summarized below:
\begin{itemize}
	
	\item We propose a deep learning based framework named NRT which can simultaneously predict precise ratings and generate abstractive tips with good linguistic quality  simulating user experience and feelings.
	All the neural parameters as well as the latent factors for users and items are learnt by a multi-task learning approach in an end-to-end training paradigm.
	
	\item We are the first to explore using tips information to improve the recommendation quality. In essence, writing some tips and giving a numerical rating are two facets of a user's product assessment action, expressing the user experience and feelings. Jointly modeling these two facets is helpful for designing a better recommendation system.
	
	\item Experimental results on benchmark datasets show that our framework achieves better performance than the state-of-the-art models on both tasks of rating prediction and abstractive tips generation.
\end{itemize}

%\section{Related Works}
\section{Related Works}
\label{section2}

Collaborative filtering (CF) has been studied for a long time and has achieved some success in recommendation systems \cite{ricci2011introduction,su2009survey}.
Latent Factor Models (LFM) based on Matrix Factorization (MF) \cite{koren2009matrix} play an important role for rating prediction. Various MF algorithms have been proposed, such as Singular Value Decomposition (SVD) and SVD++ \cite{koren2008factorization}, Non-negative Matrix Factorization (NMF) \cite{lee2001algorithms}, and Probabilistic Matrix Factorization (PMF) \cite{mnih2007probabilistic}. These methods map users and items into a shared latent factor space, and use a vector of latent features as the representation for users and items respectively. Then the inner product of their latent factor vectors can reflect the interactions between users and items.

The recommendation performance will degrade significantly when the rating matrix is very sparse. Therefore, some works consider text information for improving the rating prediction. Both item specifications and user reviews have been investigated.
In order to use the item specifications,
CTR \cite{wang2011collaborative} integrates PMF \cite{mnih2007probabilistic} and Latent Dirichlet Allocation (LDA) \cite{blei2003latent} into a single framework and employs LDA to model the text.
Collaborative Deep Learning (CDL) \cite{wang2015collaborative} employs a hierarchical Bayesian model which jointly performs deep representation learning for the specification text content and collaborative filtering for the rating matrix.
For user review texts, some research works, such as HFT \cite{mcauley2013hidden}, RMR \cite{ling2014ratings}, TriRank \cite{he2015trirank}, and sCVR \cite{rensocial2017}, integrate topic models in their frameworks to generate the latent factors for users and items incorporating review texts.
Moreover, TriRank and sCVR have been explicitly claimed that they can provide explanations for recommendations. However, one common limitation of them is that their explanations are simple extractions of words or phrases from the texts.
In contrast, we aim at generating concise sentences representing tips, which express the feeling of users while they are reviewing an item. 

Deep Learning (DL) techniques have achieved significant success in the fields of computer vision, speech recognition, and natural language processing  \cite{Goodfellow2016}. In the field of recommendation systems, researchers have made some attempts by combining different neural network structures with collaborative filtering to improve the recommendation performance.
\citet{salakhutdinov2007restricted} employ a class of two-layer Restricted Boltzmann Machines (RBM) with an efficient learning algorithm to model user interactions and perform collaborative filtering.
Considering that the training procedure of Auto-Encoders \cite{ng2011sparse} is more straightforward, some research works employ auto-encoders to tackle the latent factor modeling and rating prediction   \cite{sedhain2015autorec,wu2016collaborative,vincent2010stacked}.
%Inspired by RBM based CF and Neural Autoregressive Distribution Estimator (NADE) \cite{larochelle2011neural}, \citet{zheng2016neural} propose a feed-forward, autoregressive architecture for collaborative filtering tasks.
Recently, \citet{he2017neural} combine generalized matrix factorization and multi-layer perceptions to find better latent structures from the user interactions for improving the performance of collaborative filtering.
To model the temporal dynamic information in the user interactions, \citet{wu10recurrent} propose a recurrent recommender network which is able to predict future behavioral trajectories.
%In the industry, Google deploys a neural video recommendation system for Youtube, which can combine different types of features and be trained using large-scale datasets \cite{covington2016deep}.

%\section{Framework}
\section{Framework Description}
\label{section3}

\begin{figure*}[!t]
	\centering
	\includegraphics[width=1.6\columnwidth]{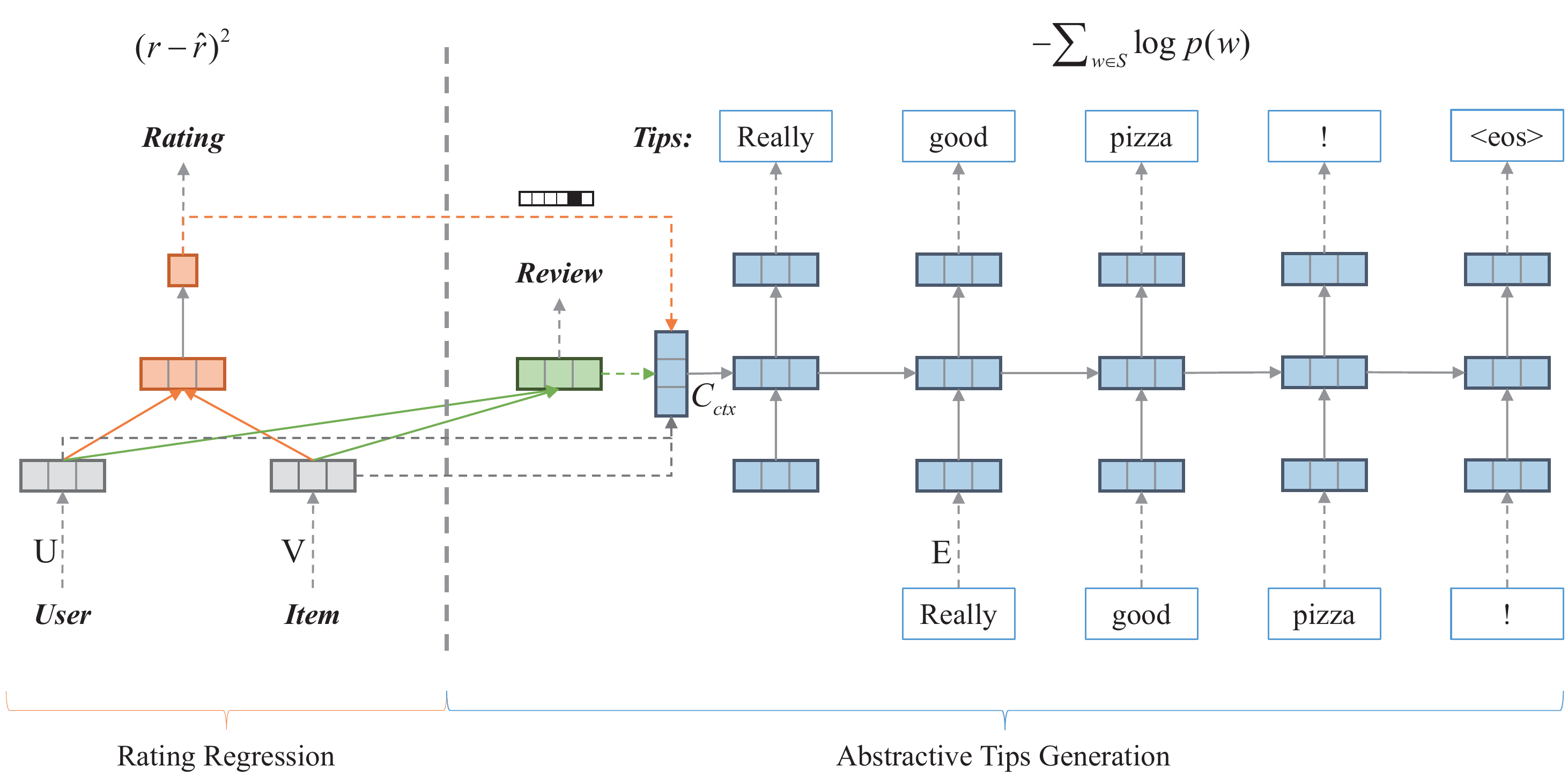}
	\caption{\label{fig:framework}
		Our proposed framework NRT for rating regression and abstractive tips generation.
	}
	\vspace{0mm}
\end{figure*}

\subsection{Overview}
The goal of recommendation, similar to collaborative filtering, is to predict a rating given a user and an item.
Additionally, in our proposed task, our model also generates abstractive tips in the form of a concise sentence. At the operational stage, only a user and an item are given.
There is no given review texts and obviously no tips texts.

At the training stage, the training data consists of users, items, tips texts, and review content.
Table~\ref{tbl:notations} depicts the notations and key concepts used in our paper.
We denote the whole training corpus by  $\mathcal{X} = \{\mathcal{U}, \mathcal{I}, \mathcal{R}, \mathcal{C}, \mathcal{S} \}$, where $\mathcal{U}$ and $\mathcal{I}$ are the sets of users and items respectively, $\mathcal{R}$ is the set of ratings, $\mathcal{C}$ is the set of review documents, and $\mathcal{S}$ is the set of tips sentences.
As shown in Figure~\ref{fig:framework}, our framework contains two major components: neural rating regression on the left and abstractive tips generation on the right.
There are two crucial latent variables
%\footnote{In different research fields, these latent variables can be called embeddings, distributed representations, or latent factors. In this work, we also call them latent factors as the others research works on recommendation systems.}
: user latent factors $\mathbf{U} \in \mathbb{R}^{k_u \times m}$ and item latent factors $\mathbf{V} \in \mathbb{R}^{k_v \times n}$, where $m$ is the number of users, and $n$ is the number of items. $k_u$ and $k_v$ are the latent factor dimension for users and items respectively.
For neural rating regression, given the user latent factor $\mathbf{u}$ and the item latent factor $\mathbf{v}$,  a multi-layer perceptron network based regression model is employed to project $\mathbf{u}$ and $\mathbf{v}$ to a real value via several layers of non-linear transformations.

For abstractive tips generation, we design a sequence decoding model based on a gated recurrent neural network called Gated Recurrent Unit (GRU) \cite{cho2014learning} to ``translate'' the combination of a user latent factor $\mathbf{u}$ and an item latent factor  $\mathbf{v}$ into a sequence of words, representing tips.
Moreover, two kinds of context information generated based on $\mathbf{u}$ and $\mathbf{v}$ are also fed into the sequence decoder model.
One is the hidden variable from the rating regression component, which is used as sentiment context information.
The other is the hidden output of a generative model for review texts.
At the operational or testing stage, we use a beam search algorithm \cite{koehn2004pharaoh} for decoding and generating the best tips given a trained model.
All the neural parameters and the latent factors for users, items, and words are  learnt by a multi-task learning approach.
The model can be trained efficiently by an end-to-end paradigm using back-propagation algorithms \cite{rumelhart1988learning}.

\begin{table}[!t]
	\caption{Glossary.}
	\label{tbl:notations}
	\centering
	\begin{tabular}{p{1.3cm} p{4cm}}
		\hline
		Symbol & Description \\
		\hline
		$\mathcal{X}$ & training set\\
		$\mathcal{V}$ & vocabulary\\
		$\mathcal{U}$  & set of users \\
		$\mathcal{I}$  & set of items \\
		$\mathcal{R}$  & set of ratings \\
		$\mathcal{C}$  & set of reviews \\
		$\mathcal{S}$  & set of tips \\
		$\mathcal{C}_{ctx}$  & context for tips decoder \\
		$\mathbf{U}$  & user latent factors \\
		$\mathbf{V}$  & item latent factors \\
		$\mathbf{E}$  & word embeddings \\
		$\mathbf{H}$  & neural hidden states \\
		$\mathbf{u}$  & user latent factor \\
		$\mathbf{v}$  & item latent factor \\
		$\mathbf{W}$  & mapping matrix\\
		$\mathbf{b}$  & bias item \\
		$\mathbf{\Theta}$  & set of neural parameters\\
		$r_{u,i}$  & rating of user $u$ to item $j$ \\
		$\sigma$ & sigmoid function \\
		$\varsigma $ & softmax function \\
		$tanh$ &  hyperbolic tangent function \\
		\hline
	\end{tabular}
	\vspace{0mm}
\end{table}

\subsection{Neural Rating Regression}
\label{sec:rating-regression}

The aim of the neural rating regression component is to conduct representation learning for the user factor $\mathbf{u}$ and the item factor $\mathbf{v}$ mentioned above.
In order to predict a rating, we need to design a model that can learn the  function $f_r(\cdot)$ which can project $\mathbf{u}$  and  $\mathbf{v}$ to a real-valued rating $\hat r$:
\begin{equation}
{\hat r} = f_r(\mathbf{u}, \mathbf{v})
\end{equation}

In most of the existing latent factor models, $f_r(\cdot)$ is represented by the inner product of $\mathbf{u}$ and $\mathbf{v}$, or adds a bias item for the corresponding user and item respectively:
\begin{equation}
\hat r = \mathbf{u}^T\mathbf{v} + b_u + b_v + b
\end{equation}
It is obvious that the rating is calculated by a linear combination of user latent factors, item latent factors, and bias.
The learnt latent factors may not capture the complex structure implied in the user historical interactions.
Recently, some research works on representation learning from different fields,  such as computer vision \cite{krizhevsky2012imagenet,goodfellow2014generative}, natural language processing \cite{mikolov2013distributed,le2014distributed}, and knowledge base completion \cite{socher2013reasoning},  demonstrate that non-linear transformations will enhance the representation ability.
Moreover, most latent factor models assume that users and items or even text information are in the same vector space and share the same latent factors.
Actually, user, item, and text information are different kinds of objects with different characteristics.
Modeling them in the same vector space would lead to limitations.

\begin{comment}
\begin{figure}[!t]
\centering
\includegraphics[width=0.9\columnwidth]{vec.pdf}
\caption{\label{fig:vec}
Uses and items are projected from their own latent vector space to a new shared space. Interactions between users and items are operated in the shared space.
}
\vspace{-3mm}
\end{figure}
\end{comment}

As shown in left part in Figure~\ref{fig:framework}, we let user latent factors  $\mathbf{U} \in \mathbb{R}^{k_u \times m}$ and item latent factors $\mathbf{V} \in \mathbb{R}^{k_v \times n}$ in different vector space, where $k_u$ and $k_v$ are the latent factor dimension for users and items respectively.
$m$ and $n$ are the number of users and items respectively.
In order to model the relationship between users and items, one may consider to use a neural tensor network \cite{socher2013reasoning} to describe the interactions between users and items, such as $\mathbf{u}^T\mathbf{W}\mathbf{v}$, where $\mathbf{W} \in \mathbb{R}^{k_u \times d \times k_v}$.
However, our investigation shows that such tensor network has too many parameters resulting in difficulty for handling large-scale datasets commonly found in recommendation applications.
Therefore, we employ a multi-layer perceptron network to model the interactions between users and items, and map user latent factors and item latent factors into real-valued ratings.

Specifically, we first map latent factors to a shared hidden space:
\begin{equation}
{\mathbf{h}^r} = \sigma (\mathbf{W}_{uh}^r\mathbf{u} + \mathbf{W}_{vh}^r\mathbf{v} + \mathbf{b}_h^r)
\label{eq:hr}
\end{equation}
where $\mathbf{W}_{uh}^r \in \mathbb{R}^{d \times k_u}$ and $\mathbf{W}_{vh}^r \in \mathbb{R}^{d \times k_v}$ are the mapping matrices for user latent factors and item latent factors respectively. $\mathbf{b}_h^r \in \mathbb{R}^{d}$ is the bias term. $d$ is the dimension of the hidden vector ${\mathbf{h}^r}$.
The superscript $r$ refers to variables related to the rating prediction component.
$\sigma(\cdot)$ is the sigmoid activation function:
\begin{equation}
\sigma (x) = \frac{1}{{1 + {e^{ - x}}}}
\end{equation}
This non-linear transformation can improve the performance of the rating prediction.
For better performance, we can add more layers of non-linear transformations into our model:
\begin{equation}
{\mathbf{h}^r_l} = \sigma (\mathbf{W}_{h{h_l}}^r\mathbf{h}^r_{l-1} + \mathbf{b}_{{h_l}}^r)
\end{equation}
where $\mathbf{W}_{h{h_l}}^r \in \mathbb{R}^{d \times d}$ is the mapping matrix for the variables in the hidden layers. $l$ is the index of a hidden layer.
Assume that $\mathbf{h}^r_L$ is the output of the last hidden layer.
The output layer transforms $\mathbf{h}^r_L$ into a real-valued rating $\hat r$:
\begin{equation}
{\hat r} = \mathbf{W}_{hr}^r\mathbf{h}^r_{L} + b^r
\label{eq:pred_r}
\end{equation}
where $\mathbf{W}_{hr}^r \in \mathbb{R}^{ d}$ and $b^r \in \mathbb{R}$.

In order to optimize the latent factors $\mathbf{U}$ and $\mathbf{V}$, as well as all the neural parameters $\Theta$,
we formulate it as a regression problem and  the loss function is formulated as:
\begin{equation}
{\mathcal{L}^r} = \frac{1}{{2\left| \mathcal{X} \right|}}{\sum\limits_{u\in \mathcal{U},i \in \mathcal{I}} {({{\hat r}_{u,i}} - {r_{u,i}})} ^2}
\label{eq:lossr}
\end{equation}
where $\mathcal{X}$ represents the training set. $r_{u,i}$ is the ground truth rating assigned by the user $u$ to the item $i$.

\subsection{Neural Abstractive Tips Generation}
Generating abstractive tips only based on user latent factors and item latent factors is a challenging task.
As mentioned above, abstractive tips generation is different from review content summarization and explainable topic words extraction.
At the operational stage, the input only consists of a user and an item, but without any text information.
After obtaining the user latent factor $\mathbf{u}$ and the item latent factor $\mathbf{v}$ from the matrices $\mathbf{U}$ and $\mathbf{V}$, we should design a strategy to ``translate'' these two latent vectors into a fluent sequence of words.
Recently, gated recurrent neural networks such as Long Short-Term Memory (LSTM) \cite{hochreiter1997long} and Gated Recurrent Unit (GRU) \cite{cho2014learning} demonstrate high capability in text generation related tasks \cite{bahdanau2014neural,rush2015neural}.
Inspired by these works and considering that GRU has comparable performance but with less parameters and more efficient computation, we employ GRU as the basic model in our sequence modeling framework. The right part of Figure~\ref{fig:framework} depicts our tips generation model.

The major idea of sequence modeling for tips generation can be expressed as follows:
\begin{equation}
p({s_t}|{s_1},{s_2}, \ldots ,{s_{t - 1}}, \mathcal{C}_{ctx}) = \varsigma (\mathbf{h}_t^s)
\end{equation}
where $s_t$ is the $t$-th word of the tips $s$.
$\mathcal{C}_{ctx}$ denotes the context information which will be described in the following sections.
$\varsigma(\cdot)$ is the softmax function and defined as follows:
\begin{equation}
\varsigma ({\mathbf{x}^{(i)}}) = \frac{{{e^{{\mathbf{x}^{(i)}}}}}}{{\sum\nolimits_{k = 1}^{K} {{e^{{\mathbf{x}^{(k)}}}}} }}
\label{eq:sfotmax}
\end{equation}
$\mathbf{h}_t^s$ is the sequence hidden state at the time $t$ and it depends on the input at the time $t$ and the previous hidden state  $\mathbf{h}_{t-1}^s$:
\begin{equation}
\mathbf{h}_{t}^s =  f(\mathbf{h}_{t-1}^s, s_{t})
\end{equation}
Here $f(\cdot)$ can be the vanilla RNN, LSTM, or GRU.
In the case of GRU, the state updates are processed according to the following operations:
\begin{equation}
\begin{array}{l}
\mathbf{r}_t^s = \sigma (\mathbf{W}_{sr}^s\mathbf{s}_t + \mathbf{W}_{hr}^s\mathbf{h}_{t - 1}^s + \mathbf{b}_r^s)\\
\mathbf{z}_t^s = \sigma (\mathbf{W}_{sz}^s\mathbf{s}_t + \mathbf{W}_{hz}^s\mathbf{h}_{t - 1}^s + \mathbf{b}_z^s)\\
\mathbf{g}_t^s = \tanh (\mathbf{W}_{sh}^s\mathbf{s}_t + \mathbf{W}_{hh}^s(\mathbf{r}_t^s \odot \mathbf{h}_{t - 1}^s) + \mathbf{b}_h^s)\\
\mathbf{h}_t^s = \mathbf{z}_t^s \odot \mathbf{h}_{t - 1}^s + (1 - \mathbf{z}_t^s) \odot \mathbf{g}_t^s
\end{array}
\label{eq:gru}
\end{equation}
where $\mathbf{s}_t \in \mathbf{E}$ is the embedding vector for the word $s_t$ of the tips and the vector is also learnt from our framework.
$\mathbf{r}_t^s$ is the reset gate, $\mathbf{z}_t^s$ is the update gate.
$\odot$ denotes element-wise multiplication. $tanh$ is the  hyperbolic tangent activation function.

As shown in Figure~\ref{fig:framework}, when $t=1$, the sequence model has no input information.
Therefore, we utilize the context information $\mathcal{C}_{ctx}$ to initialize $\mathbf{h}_{0}^s$.
Context information is very crucial in a sequence decoding framework, which will directly affect the performance of sequence generation.
In the field of neural machine translation \cite{wu2016google}, context information includes the encoding information of the source input and the decoding attention information from the source.
In the field of neural summarization \cite{rush2015neural,li2017salience}, the context is the encoded document information.
In our framework, the corresponding user $u$ and item $i$ are the input from which we design two kinds of context information for tips generation:
predicted rating $\hat r_{u,i}$ and the generated hidden variable for the review text $\mathbf{h}^c_L$.

For the input, we just find the user latent factor and the item latent factor from the matrices $\mathbf{U}$ and $\mathbf{V}$:
\begin{equation}
\mathbf{u}=\mathbf{U}(:,u), \mathbf{v}=\mathbf{V}(:,i)
\end{equation}

For the context of rating information, we can employ the output of the rating regression component in Section~\ref{sec:rating-regression}.
Specifically, after getting the predicted rating $\hat r_{u,i}$, for example, $\hat r_{u,i} = 4.321$, we cast it into an integer $4$, and add a step of vectorization. Then we get the vector representation of rating $\hat r_{u,i}$. If the rating range is $[0, 5]$, we will get the rating vector $ \mathbf{\hat r}_{u,i}$:
\begin{equation}
\mathbf{\hat r}_{u,i} = (0, 0, 0, 0, 1, 0)^T
\end{equation}
$ \mathbf{\hat r}_{u,i}$ is used as the context information to control the sentiment of the generated tips.

Another context information is from review texts.
One should note that review texts cannot be used as the input directly.
The reason is that at the testing state, there are no review information.
We only make use of reviews to enhance the representation ability of the latent vectors $\mathbf{U}$ and  $\mathbf{V}$.
We develop a standard generative model for review texts based on a multi-layer perceptron. For review content $c_{u,i}$ written by the user $u$ to the item $i$, the generative process is defined as follows.
We first map the user latent vector $\mathbf{u}$ and the item latent factor  $\mathbf{v}$ into a hidden space:
\begin{equation}
{\mathbf{h}^c} = \sigma (\mathbf{W}_{uh}^c\mathbf{u} + \mathbf{W}_{vh}^c\mathbf{v} + \mathbf{b}_h^c)
\label{eq:hc}
\end{equation}
It is obvious that we can also add more layers of non-linear transformation into the generative hidden layers.
Assume that $\mathbf{h}^c_L$ is the output of the last hidden layer.
We add the final generative layer to map $\mathbf{h}^c_L$ into a $|\mathcal{V}|$-size vector $\mathbf{\hat c}$, where $\mathcal{V}$ is the vocabulary of words in the reviews and the tips:
\begin{equation}
{\mathbf{\hat c}} = \varsigma(\mathbf{W}_{hc}^c\mathbf{h}^c_{L} + \mathbf{b}^c)
\end{equation}
where $\mathbf{W}_{hc}^c \in \mathbb{R}^{|\mathcal{V}| \times d}$ and $\mathbf{b}^c \in \mathbb{R}^{|\mathcal{V}|}$. $\varsigma(\cdot)$ is the softmax function defined in Equation~\ref{eq:sfotmax}.
In fact we can regard $\mathbf{\hat c}$ as a multinomial distribution defined on $\mathcal{V}$. Therefore, we can draw some words from $\mathbf{\hat c}$ and generate the content of the review $c_{u,i}$.
We let $\mathbf{c}$ be the ground truth of $c_{u,i}$.
$\mathbf{c}^{(k)}$ is the term frequency of the word $k$ in $c_{u,i}$.
We employ the likelihood to evaluate the performance of this generative process.
For convenience, we use the Negative Log-Likelihood (NLL) as the loss function:
\begin{equation}
{\mathcal{L}^c} =  - \sum\limits_{k = 1}^{|\mathcal{V}|} {{\mathbf{c}^{(k)}}\log {\mathbf{\hat c}^{(k)}}}
\label{eq:lossc}
\end{equation}

One characteristic of the design of our model is that both the rating and review texts are generated from the same user latent factors $\mathbf{U}$ and item latent factors $\mathbf{V}$, i.e., $\mathbf{U}$ and $\mathbf{V}$ are shared by the subtasks of rating prediction and review text generation.
Thus, in the training stage, both of $\mathbf{U}$ and $\mathbf{V}$ receive the feedback from all the subtasks, which improves the representation ability of the latent factors.

After obtaining all the context information $\mathcal{C}_{ctx} = \{ \mathbf{\hat r}, \mathbf{h}_L^c\}$, we integrate them into the initial decoding hidden state $\mathbf{h}_{0}^s$ using a non-linear transformation:
\begin{equation}
\mathbf{h}_{0}^s = tanh(\mathbf{W}_{uh}^s\mathbf{u} + \mathbf{W}_{vh}^s\mathbf{v} + \mathbf{W}_{rh}^s\mathbf{\hat r} + \mathbf{W}_{ch}^s\mathbf{h}_L^c + \mathbf{b}_c^s)
\end{equation}
where $\mathbf{u}$ is the user latent factor, $\mathbf{v}$ is the item latent factor, $\mathbf{\hat r}$ is the vectorization for the predicted rating $\hat r$, and $\mathbf{h}_L^c$ is the generated hidden variable from the review text.
Then GRU can conduct the sequence decoding progress. After getting all the sequence hidden states, we feed them to the final output layer to predict the word sequence in tips.
\begin{equation}
\mathbf{\hat s}_{t+1} = \varsigma (\mathbf{W}_{hs}^s\mathbf{h}_t^s  + \mathbf{b}^s)
\end{equation}
where $\mathbf{W}_{hs}^s \in \mathbb{R}^{d \times |\mathcal{V}|}$ and $\mathbf{b}^s \in \mathbb{R}^{|\mathcal{V}|}$. $\varsigma(\cdot)$ is the softmax function defined in Equation~\ref{eq:sfotmax}. Then the word with the largest probability is the decoding result for the step $t+1$:
\begin{equation}
{w_{t + 1}^*} = \mathop {\arg\max }\limits_{{w_i} \in \mathcal{V}} \mathbf{\hat s}_{t + 1}^{ (w_i) }
\end{equation}

At the training stage, we also use NLL as the loss function, where $I_w$ is the vocabulary index of the  word $w$:
\begin{equation}
{\mathcal{L}^s} =  - \sum\limits_{w \in Tips} {\log {\mathbf{\hat s}^{(I_w)}}}
\label{eq:losss}
\end{equation}

At the testing stage, given a trained model, we employ the beam search algorithm to find the best sequence $s^*$ having the maximum log-likelihood.
\begin{equation}
{s^*} = \mathop {\arg \max }\limits_{s \in \mathcal{S}} \sum\limits_{w \in s} {\log {\mathbf{\hat s}^{(I_w)}}}
\end{equation}

The details of the beam search algorithm is shown in Algorithm~\ref{alg:bs}.

\begin{algorithm}[!t]
	\caption{Beam search for abstractive tips generation}
	\label{alg:bs}
	\begin{algorithmic}[1]
		\REQUIRE Beam size $\beta$, maximum length $\eta$, user id $u$, item id $v$, and tips generation model $\mathcal{G}$.
		\ENSURE $\beta$ best candidate tips.
		\STATE Initialize $\Pi = \emptyset$, $\pi[0:\beta-1] = \mathbf{0}$,  $\Pi{_p} = \emptyset $,  $\pi_p = \mathbf{0}$, $t = 0$;
		\STATE Get user latent factor and item latent factor:\\
		\ \ \ \ $\mathbf{u}=\mathbf{U}(:,u)$ and $ \mathbf{v}=\mathbf{V}(:,v)$
		\WHILE{$t < \eta$ }
		\STATE Generate $\beta$ new states based on $\Pi$: $\{ \mathbf{\hat s}_t\}_0^{\beta-1}$ = $\mathcal{G}(\Pi)$
		\FOR{$i$ from $0$ to $\beta$}
		\STATE Uncompleted sequence $s_i \leftarrow \Pi(i)$
		\STATE Top-$\beta$ words  $\{w_0, w_1, \ldots, w_{\beta-1}\} \leftarrow \beta\textrm{-}\mathop {\arg\max}\limits_{{w_i} \in \mathcal{V}} \mathbf{\hat s}_{t_i}^{ (w_i) }$
		\FOR{each word $w_j$}
		\STATE Concatenation: $\Pi_p.inseart(s_i + w_j) $
		\STATE Likelihood: $\pi_p.inseart(\pi[i] + \log\mathbf{\hat s}_{t_i}^{ (w_j) })$
		\ENDFOR
		\ENDFOR
		\STATE Get the top-$\beta$ sequences with largest likelihood:\\
		\ \ \ \ \ \ \  \ $\{s\}_0^{\beta-1}, \{l\}_0^{\beta-1} = \beta\textrm{-}\mathop{\arg \max }\limits_{{s} \in \Pi_p, l \in \pi_p} l$
		\STATE $\Pi \leftarrow\{s\}_0^{\beta-1}$, $\pi \leftarrow \{l\}_0^{\beta-1}$, $\Pi{_p} = \emptyset $,  $\pi_p = \mathbf{0}$
		\STATE $t \leftarrow t + 1$
		\ENDWHILE
		\RETURN $\Pi$, $\pi$.
	\end{algorithmic}
\end{algorithm}

\subsection{Multi-task Learning}
We integrate all the subtasks of rating prediction and abstractive tips generation into a unified multi-task learning framework whose objective function is:
\begin{equation}
{\mathcal{J}} = \mathop {\min }\limits_{\mathbf{U}, \mathbf{V}, \mathbf{E}, \Theta} (\lambda_r\mathcal{L}^r + \lambda_c\mathcal{L}^c + \lambda_s{\mathcal{L}^s} + \lambda_n(\left\| \mathbf{U}  \right\|_2^2 + \left\| \mathbf{V}  \right\|_2^2 + \left\| \Theta  \right\|_2^2))
\label{eq:obj}
\end{equation}
where $\mathcal{L}^r$ is the rating regression loss from Equation~\ref{eq:lossr},
$\mathcal{L}^c$ is the review text generation loss from Equation~\ref{eq:lossc},
and $\mathcal{L}^r$ is the tips generation loss from Equation~\ref{eq:losss}.
$\Theta$ is the set of neural parameters.
$\lambda_r$, $\lambda_c$, $\lambda_s$, and $\lambda_n$ are the weight proportion of each term.
The whole framework can be efficiently trained using back-propagation in an end-to-end paradigm.

%\section{Experiments}

\section{Experimental Setup}
\label{section4}

\subsection{Research Questions}
We list the research questions we want to investigate in this paper:

\begin{itemize}
	\item \textbf{RQ1}: What is the performance of NRT in rating prediction tasks? Does it outperform the state-of-the-art models? (See Section~\ref{sec:exp:rating}.)
	
	\item \textbf{RQ2}: What is the performance of NRT in abstractive tips generation? Can the generated tips express user experience and feelings? (See Section~\ref{sec:exp:tips})
	
	\item \textbf{RQ3}: What is the relationship between predicted ratings and the sentiment of generated tips? (See Section~\ref{sec:exp:case})
	
	%\item \textbf{RQ4}: Performance of embeddings for users and items? (See Section~\ref{sec:exp:case})
\end{itemize}

We conduct extensive experiments to investigate the above research questions.

\subsection{Datasets}
In our experiments, we use four standard benchmark datasets from different domains to evaluate our model. The ratings of these datasets are integers in the range of $[0, 5]$.
There are three datasets from Amazon 5-core\footnote{http://jmcauley.ucsd.edu/data/amazon}: \textbf{Books}, \textbf{Electronics}, and \textbf{Movies \& TV}.
``Books'' is the largest dataset among all the domains. It contains 603,668 users, 367,982 items, and  8,887,781 reviews.
%``Electronics'' contains 192,403 users, 63,001 items, and 1,684,779 reviews.
%``Movies\&TV'' contains 123,960 users, 50,052 items, and 1,697,533 reviews.
We regard the field ``summary'' as tips, and the number of tips texts is same with the number of reviews.

Another dataset is from \textbf{Yelp} Challenge 2016\footnote{https://www.yelp.com/dataset\_challenge}.
It is also a large-scale dataset consisting of restaurant reviews and tips.
%It contains 85,533 items and 2,346,227 reviews.
The number of users is 684,295, which is the largest among all the datasets. Therefore this dataset is also the most sparse one. Tips are included in the dataset. For samples without tips, the first sentence of review texts is extracted and regarded as tips.

We filter out the words with low term frequency in the tips and review texts, and build a vocabulary $\mathcal{V}$ for each dataset. We show the statistics of our datasets in Table~\ref{tbl:rec_dataset}.

\begin{table}[!t]
	\centering
	\caption[datasets]{Overview of the datasets.}
	\label{tbl:rec_dataset}
	\begin{tabular}{lrrrr}
		\toprule
		& \textbf{Books} & \textbf{Electronics} &\textbf{ Movies\&TV} & \textbf{Yelp-2016}  \\
		\midrule
		\textit{users}     & 603,668  & 192,403 & 123,960  & 684,295 \\
		\textit{items} & 367,982  & 63,001  & 50,052  & 85,533 \\
		\textit{reviews}   & 8,887,781  & 1,684,779 & 1,697,533   & 2,346,227\\
		$|\mathcal{V}|$   &  258,190 &  70,294 & 119,530   &  111,102 \\
		\bottomrule
	\end{tabular}
	\vspace{0mm}
\end{table}

\subsection{Evaluation Metrics}
For the evaluation of rating prediction,
we employ two metrics: Mean Absolute Error (\textit{MAE}) and Root
Mean Square Error (\textit{RMSE}).
Both of them are widely used for rating prediction in recommender systems. Given a predicted rating $\hat r_{u,i}$ and a ground-truth rating $r_{u,i}$ from the user $u$ for the item $i$, the RMSE is calculated as:
\begin{equation}\label{eq:expst1}
RMSE = \sqrt {\frac{1}{N}\sum\limits_{u,i} {{{({r_{u,i}} - \hat r_{u,i})}^2}} }
\end{equation}
where $N$ indicates the number of ratings between users and items. Similarly, MAE is calculated as follows:
\begin{equation}\label{eq:expst2}
MAE = \frac{1}{N}\sum\limits_{u,i} {\left| {{r_{u,i}} - \hat r_{u,i}} \right|}
\end{equation}

For the evaluation of abstractive tips generation, the ground truth $s_h$ is the
tips written by the user for the item.
We use \textit{ROUGE} \cite{lin2004rouge} as our evaluation metric with standard options\footnote{ROUGE-1.5.5.pl -n 4 -w 1.2 -m -2 4 -u -c 95 -r 1000 -f A -p 0.5 -t 0}. It is a classical evaluation metric in the field of text summarization \cite{lin2004rouge,lidong15absmds}.
It counts the number of overlapping units between the generated tips and the ground truth written by users. Assuming that $s$ is the generated tips, $g_n$ is n-gram, ${C}({g_n})$ is the number of n-grams in $\tilde s$ ($s_h$ or $s$), ${C_m}({g_n})$ is the number of n-grams co-occurring in $s$ and $s_h$, then the ROUGE-N score for $s$ is defined as follows:
\begin{equation}
{ROUGE\textrm{-}N}(s) = \sum\nolimits_{{g_n} \in {s_h}} {{C_m}({g_n})/\sum\nolimits_{{g_n} \in {\tilde s}} {C({g_n})} }
\end{equation}
When $\tilde s = s_h$, we can get $ROUGE_{recall}$, and when $\tilde s = s$, we get $ROUGE_{presicion}$.  We use Recall, Precision, and F-measure of ROUGE-1 (R-1), ROUGE-2 (R-2), ROUGE-L (R-L), and ROUGE-SU4 (R-SU4) to evaluate the quality of the generated tips.

\subsection{Comparative Methods}

To evaluate the performance of rating prediction, we compare our model with the following methods:

\begin{itemize}
	
	\item \textbf{RMR}: \textbf{R}atings \textbf{M}eet \textbf{R}eviews \cite{ling2014ratings}. %It proposes a unified framework which combines content-based filtering with collaborative filtering.
	It utilizes a topic modeling technique to model the review texts and achieves significant improvements compared with other strong topic modeling based methods.
	
	\item \textbf{CTR}: \textbf{C}ollaborative \textbf{T}opic \textbf{R}egression \cite{wang2011collaborative}. It is a popular method for scientific articles recommendation by solving a one-class collaborative filtering problem. Note that CTR uses both ratings and item specifications.

	\item \textbf{NMF}: \textbf{N}on-negative \textbf{M}atrix \textbf{F}actorization \cite{lee2001algorithms}.
	%The non-negativity constraints are integrated in the typical matrix factorization and make the representation purely additive.
	%NMF is a popular and strong baseline for CF-based recommendation.
	It only uses the rating matrix as the input.
	
	\item \textbf{PMF}: \textbf{P}robabilistic \textbf{M}atrix \textbf{F}actorization \cite{mnih2007probabilistic}.
	Gaussian distribution is introduced to model the latent factors for users and items.
	
	\item \textbf{LRMF}: \textbf{L}earning to \textbf{R}ank with \textbf{M}atrix \textbf{F}actorization \cite{shi2010list}. It combines a list-wise learning-to-rank algorithm with matrix factorization to improve recommendation.
	
	\item \textbf{SVD++}: It extends \textbf{S}ingular \textbf{V}alue \textbf{D}ecomposition by considering implicit feedback information for latent factor modeling \cite{koren2008factorization}.
	
	\item \textbf{URP}: \textbf{U}ser \textbf{R}ating \textbf{P}rofile modeling \cite{marlin2003modeling}. Topic models are employed to model the user preference from a generative perspective. It still only uses the rating matrix as input.
	
\end{itemize}

For abstractive tips generation, we find that no existing works can generate abstractive tips purely based on latent factors of users and items.
In order to evaluate the performance and conduct comparison with some baselines, we refine some existing methods to make them capable of extracting sentences for tips generation as follows.

\textbf{LexRank} \cite{erkan2004lexrank} is a classical method in the field of text summarization.
We add a preprocessing procedure to prepare the input texts for LexRank, which consists of the following steps: (1) Retrieval: For the user $u$, we first retrieve all her reviews $\mathcal{C}_u$ from the training set. For the item $i$, we use the same method to get $\mathcal{C}_i$. (2) Filtering: Assuming that the ground truth rating for $u$ and $i$ is $r_{u,i}$, then we remove all the reviews from $\mathcal{C}_u$ and $\mathcal{C}_i$ whose ratings are not equal to $r_{u,i}$. The reviews whose words only appear in one set are also removed.
(3) Tips extraction: We first merge $\mathcal{C}_u$ and $\mathcal{C}_i$ to get $\mathcal{C}_{u,i}$, then the problem can be regarded as a multi-document summarization problem. LexRank can extract a sentence from $\mathcal{C}_{u,i}$ as the final tips. Note that we give an advantage of this method since the ground truth ratings are used.

CTR contains a topic model component and it can generate topics for items. So the topic related variables are employed to extract tips: (1) We first get the latent factor $\theta_i$ for item $i$, and draw the topic $z$ with the largest probability from $\theta_i$. Then from $\phi_z$, which is a multinomial distribution of $z$ on $\mathcal{V}$, we select the top-$50$ words with the largest probability. (2) The most similar sentence from $\mathcal{C}_{u,i}$ is extracted as the tips. This baseline is named \textbf{CTR$_t$}.
Another baseline method \textbf{RMR$_t$} is designed in the same way.

Finally, we list all the methods and baselines in Table~\ref{tbl:baselinesandmethods}.

\begin{table}[t]
	\centering
	%\small
	\caption{Baselines and methods used for comparison.}
	\label{tbl:baselinesandmethods}
	\begin{tabular}{l l r}
		\toprule
		Acronym & Gloss & \mbox{}\hspace*{-1cm} Reference \\
		\midrule
		NRT & Neural rating and tips generation & \mbox{}\hspace*{-1cm}Section~\ref{section3}\\
		
		\midrule
		\multicolumn{3}{@{}l}{\emph{Rating prediction}}\\
		RMR & Ratings meet reviews model & \cite{ling2014ratings} \\
		CTR & Collaborative topic regression model  & \cite{wang2011collaborative} \\
		NMF & Non-negative matrix factorization & \cite{lee2001algorithms} \\
		PMF & Probabilistic matrix factorization & \cite{mnih2007probabilistic} \\
		LRMF & List-wise learning to rank for item ranking & \cite{shi2010list} \\
		SVD++ & Factorization meets the neighborhood & \cite{koren2008factorization} \\
		URP & User rating profile modeling using LDA & \cite{marlin2003modeling} \\
		%HFT & Hidden factors as topics model & \cite{mcauley2013hidden} \\
		
		\multicolumn{3}{@{}l}{\emph{Tips generation}}\\
		LexRank & Pagerank for summarization& \cite{erkan2004lexrank} \\
		CTR$_t$ & CTR for tips topic extraction & \cite{wang2011collaborative} \\
		RMR$_t$ & RMR for tips topic extraction & \cite{ling2014ratings} \\
		\bottomrule
	\end{tabular}
	\vspace{0mm}
\end{table}

\subsection{Experimental Settings}
Each dataset is divided into three subsets: $80\%$, $10\%$, and $10\%$, for training, validation, and testing, receptively. All the parameters of our model  are tuned with the validation set.
After the tuning process, we set the number of latent factors $k = 10$ for LRMF, NMF, PMF, and SVD++.
We set the number of topics $K = 50$ for the methods using topic models.
In our model NRT, we set $K = 300$ for user latent factors, item latent factors, and word latent factors. The dimension of the hidden size is $400$.
The number of layers for the rating regression model is $4$, and for the tips generation model is $1$.
We set the beam size $\beta = 4$, and the maximum length $\eta = 20$.
For the optimization objective, we let the weight parameters $\lambda_r = \lambda_c = \lambda_s = 1$, and $\lambda_n = 0.0001$.
The batch size for mini-batch training is $200$.

All the neural matrix parameters in hidden layers and RNN layers are initialized from a uniform distribution between $[-0.1, 0.1]$.
Adadelta \cite{zeiler2012adadelta} is used for gradient based optimization.
Our framework is implemented with Theano \cite{2016arXiv160502688short} on a single Tesla K80 GPU.

\section{Results and Discussions}
\label{section5}

\subsection{Rating Prediction (RQ1)}
\label{sec:exp:rating}

\begin{table*}[t]
	\begin{threeparttable}
		\centering
		\caption{ \textbf{MAE} and \textbf{RMSE} values for rating prediction. }
		\label{tab:rmse}
		\begin{tabular}{@{}lcc cc cc cc@{}}
			\toprule
			& \multicolumn{2}{c}{Books} & \multicolumn{2}{c}{Electronics} &
			\multicolumn{2}{c}{Movies} & \multicolumn{2}{c}{Yelp-2016}\\
			\cmidrule(lr){2-3}
			\cmidrule(lr){4-5}
			\cmidrule(lr){6-7}
			\cmidrule(lr){8-9}
			& MAE & RMSE  & MAE & RMSE & MAE & RMSE & MAE & RMSE\\
			\midrule
			%CliMF & -\phantom{0} &-\phantom{0} & -\phantom{0} & -\phantom{0}  & -\phantom{0} & -\phantom{0}   & -\phantom{0} & -\phantom{0} \\
			LRMF & 1.939\phantom{0} & 2.153\phantom{0}
			& 2.005\phantom{0} & 2.203\phantom{0}
			& 1.977\phantom{0} & 2.189\phantom{0}
			& 1.809\phantom{0} & 2.038\phantom{0}   \\
			PMF & 0.882\phantom{0} & 1.219\phantom{0}
			& 1.220\phantom{0} & 1.612\phantom{0}
			& 0.927\phantom{0} & 1.290\phantom{0}
			& 1.320\phantom{0} & 1.752\phantom{0} \\
			NMF & 0.731\phantom{0} & 1.035\phantom{0}
			& 0.904\phantom{0} & 1.297\phantom{0}
			& 0.794\phantom{0} & 1.135\phantom{0}
			& 1.062\phantom{0} & 1.454\phantom{0}  \\
			SVD++ & 0.686\phantom{0} & 0.967\phantom{0}
			& 0.847\phantom{0} & 1.194\phantom{0}
			& 0.745\phantom{0} & 1.049\phantom{0}
			& 1.020\phantom{0} & 1.349\phantom{0} \\
			URP & 0.704\phantom{0} & 0.945\phantom{0}
			& 0.860\phantom{0} & 1.126\phantom{0}
			& 0.764\phantom{0} & 1.006\phantom{0}
			& 1.030\phantom{0} & 1.286\phantom{0} \\
			CTR &  0.736\phantom{0} &  0.961\phantom{0}
			&  0.903\phantom{0} &  1.154\phantom{0}
			& 0.854\phantom{0} &  1.069\phantom{0}
			& 1.174\phantom{0} &  1.392\phantom{0} \\
			RMR & 0.681\phantom{0} & 0.933\phantom{0}
			& 0.822\phantom{0} & 1.123\phantom{0}
			& 0.741\phantom{0} & 1.005\phantom{0}
			& 0.994\phantom{0} &  1.286\phantom{0} \\
			\textbf{NRT} & \textbf{0.667}\textbf{*} & \textbf{0.927}\textbf{*}
			& \textbf{0.806}\textbf{*} & \textbf{1.107}\textbf{*}
			& \textbf{0.702}\textbf{*} & \textbf{0.985}\textbf{*}
			& \textbf{0.985}\textbf{*} & \textbf{1.277}\textbf{*}  \\
			\bottomrule
		\end{tabular}
		\begin{tablenotes}
			\small
			\item \textbf{*}Statistical significance tests show that our method is better than RMR \cite{ling2014ratings}.
		\end{tablenotes}
	\end{threeparttable}
\end{table*}

The rating prediction results of our framework NRT and comparative models on all datasets are given in Table~\ref{tab:rmse}.
It shows that our model consistently outperforms all comparative methods under both MAE and RMSE
metrics on all datasets.
From the comparison,  we notice that the topic modeling based methods CTR and RMR are much better than LRMF, NMF, PMF, and SVD++.
The reason is that CTR and RMR consider text information such as item specifications and user reviews to improve the representation quality of latent factors,
while the traditional CF-based models (e.g. LRMF, NMF, PMF, and SVD++) only consider the rating matrix as the input.
Statistical significance of differences between the performance of NRT and RMR, the best comparison method, is tested using a two-tailed paired t-test. The result shows that NRT is significantly better than RMR.

Except jointly learning the tips decoder, we did not apply any sophisticated linguistic operations on the texts of reviews and tips. Jointly modeling the tips information is already very helpful for recommendation performance.
In fact, tips and its corresponding rating are two facets of
product assessment by a user on an item, namely, the qualitative facet and the quantitative facet.
Our framework NRT elegantly captures this information with its multi-task learning model. Therefore the learnt latent factors are more effective.

\subsection{Abstractive Tips Generation (RQ2)}
\label{sec:exp:tips}

Our NRT model can not only solve the rating prediction problem, but also generate abstractive tips simulating how users express their experience and feelings.
The evaluation results of tips generation of our model and the comparative methods are given in Table~\ref{tab:rouge-books}$\sim$Table~\ref{tab:rouge-yelp}.
In order to capture more details, we report Recall, Precision, and F-measure (in percentage) of ROUGE-1, ROUGE-2, ROUGE-L, and ROUGE-SU4.
Our model achieves the best performance in the metrics of Precision and F1-measure among all the four datasets.
On the dataset of Movies\&TV, NRT also achieves the best Recall for all ROUGE metrics.

\begin{table*}[!htb]
	\centering
	\caption{ROUGE evaluation on dataset Books.}
	\label{tab:rouge-books}
	\begin{tabular}{|c|c|c|c|c|c|c|c|c|c|c|c|c|}
		\hline
		\multirow{2}{*}{\textbf{Methods}} &
		\multicolumn{3}{c|}{\textbf{ROUGE-1}} &
		\multicolumn{3}{c|}{\textbf{ROUGE-2}} &
		\multicolumn{3}{c|}{\textbf{ROUGE-L}} &
		\multicolumn{3}{c|}{\textbf{ROUGE-SU4}} \\
		\cline{2-13}
		& R & P & F1 & R & P & F1& R & P & F1& R & P & F1 \\
		%\hline
		%LexRank-T & 16.48 & 16.28 & 15.32
		%& 7.53 & 7.65 & 7.39
		%& 8.75 & 6.58 & 6.85
		%& 4.63 & 3.62 & 3.51 \\
		\hline
		LexRank & 12.94 & 12.02 & 12.18
		& 2.26 & 2.29 & 2.23
		& 11.72 & 10.89 & 11.02
		& 4.13 & 4.15 & 4.02 \\
		%\hline
		%RMR-T & 18.53 & 18.82 & 14.86
		%& 4.24 & 3.73 & 3.83
		%& 16.08 & 11.96 & 12.87
		%& 7.41 & 5.59 & 5.84 \\
		\hline
		RMR$_t$ & 13.80 & 11.69 & 12.43
		& 1.79 & 1.57 & 1.64
		& 12.54 & 10.55 & 11.25
		& 4.49 & 3.54 & 3.80 \\
		%\hline
		%CTR-T & 18.65 & 13.84 & 14.93
		%& 4.24 & 3.70 & 3.81
		%& 16.19 & 11.97 & 12.92
		%& 7.43 & 5.59 & 5.84 \\
		\hline
		CTR$_t$ & 14.06 & 11.85 & 12.62
		& 2.03 & 1.80 & 1.87
		& 12.68 & 10.64 & 11.35
		& 4.71 & 3.71 & 3.99 \\
		\hline
		\textbf{NRT} & 10.30 & \textbf{19.28} & \textbf{12.67 }
		& 1.91 & \textbf{3.76} & \textbf{2.36}
		& 9.71 & \textbf{17.92} & \textbf{11.88}
		& 3.24 & \textbf{8.03} & \textbf{4.13} \\
		\hline
		
	\end{tabular}
	\vspace{0mm}
\end{table*}

\begin{table*}[!htb]
	\centering
	
	\caption{ROUGE evaluation on dataset Electronics.}
	\label{tab:rouge-elects}
	\begin{tabular}{|c|c|c|c|c|c|c|c|c|c|c|c|c|}
		\hline
		\multirow{2}{*}{\textbf{Methods}} &
		\multicolumn{3}{c|}{\textbf{ROUGE-1}} &
		\multicolumn{3}{c|}{\textbf{ROUGE-2}} &
		\multicolumn{3}{c|}{\textbf{ROUGE-L}} &
		\multicolumn{3}{c|}{\textbf{ROUGE-SU4}} \\
		\cline{2-13}
		& R & P & F1 & R & P & F1& R & P & F1& R & P & F1 \\
		%\hline
		%LexRank-T & 18.54 & 19.04 & 17.70
		%& 9.93 & 10.40 & 9.90
		%& 17.41 & 17.69 & 16.61
		%& 11.77 & 12.13 & 11.29 \\
		\hline
		LexRank & 13.42 & 13.48 & 12.08
		& 1.90 & 2.04 & 1.83
		& 11.72 & 11.48 & 10.44
		& 4.57 & 4.51 & 3.88 \\
		%\hline
		%RMR-T & 16.42 & 13.66 & 13.86
		%& 4.49 & 4.04 & 4.11
		%& 14.53 & 12.13 & 12.30
		%& 7.04 & 5.92 & 5.89 \\
		\hline
		RMR$_t$ & 15.68 & 11.32 & 12.30
		& 2.52 & 2.04 & 2.15
		& 13.37 & 9.61 & 10.45
		& 5.41 & 3.72 & 3.97 \\
		%\hline
		%CTR-T & 16.06 & 13.36 & 13.51
		%& 4.02 & 3.71 & 3.72
		%& 14.05 & 11.76 & 11.86
		%& 6.68 & 5.61 & 5.54 \\
		\hline
		CTR$_t$ & 15.81 & 11.37 & 12.38
		& 2.49 & 1.92 & 2.05
		& 13.45 & 9.62 & 10.50
		& 5.39 & 3.63 & 3.89 \\
		\hline
		\textbf{NRT} & 13.08 & \textbf{17.72} & \textbf{13.95}
		& \textbf{2.59} & \textbf{3.36} & \textbf{2.72}
		& 11.93 & \textbf{16.01} & \textbf{12.67}
		& 4.51 & \textbf{6.69} & \textbf{4.68} \\
		\hline
	\end{tabular}
	\vspace{0mm}
\end{table*}

\begin{table*}[!htb]
	\centering
	
	\caption{ROUGE evaluation on dataset Movies\&TV.}
	\label{tab:rouge-movies}
	\begin{tabular}{|c|c|c|c|c|c|c|c|c|c|c|c|c|}
		\hline
		\multirow{2}{*}{\textbf{Methods}} &
		\multicolumn{3}{c|}{\textbf{ROUGE-1}} &
		\multicolumn{3}{c|}{\textbf{ROUGE-2}} &
		\multicolumn{3}{c|}{\textbf{ROUGE-L}} &
		\multicolumn{3}{c|}{\textbf{ROUGE-SU4}} \\
		\cline{2-13}
		& R & P & F1 & R & P & F1& R & P & F1& R & P & F1 \\
		%\hline
		%LexRank-T & 13.78 & 14.05 & 12.79
		%& 4.20 & 4.11 & 3.95
		%& 12.65 & 12.74 & 11.70
		%& 7.20 & 7.61 & 6.70 \\
		\hline
		LexRank & 13.62 & 14.11 & 12.37
		& 1.92 & 2.09 & 1.81
		& 11.69 & 11.74 & 10.47
		& 4.47 & 4.53 & 3.75 \\
		%%\hline
		%RMR-T & 14.66 & 11.31 & 11.90
		%& 2.41 & 2.09 & 2.13
		%& 12.70 & 9.70 & 10.24
		%& 5.08 & 3.79 & 3.89 \\
		\hline
		RMR$_t$ & 14.64 & 10.26 & 11.33
		& 1.78 & 1.36 & 1.46
		& 12.62 & 8.72 & 9.67
		& 4.63 & 3.00 & 3.28 \\
		%\hline
		%CTR-T & 14.94 & 11.26 & 11.95
		%& 2.46 & 2.03 & 2.11
		%& 13.02 & 9.76 & 10.37
		%& 5.31 & 3.83 & 3.97 \\
		\hline
		CTR$_t$ & 15.13 & 10.37 & 11.57
		& 1.90 & 1.42 & 1.54
		& 13.02 & 8.77 & 9.85
		& 4.88 & 3.03 & 3.36 \\
		\hline
		\textbf{NRT} & \textbf{15.17} & \textbf{20.22} & \textbf{16.20}
		&\textbf{ 4.25} & \textbf{5.72} &\textbf{ 4.56}
		& \textbf{13.82} & \textbf{18.36} & \textbf{14.73}
		& \textbf{6.04} & \textbf{8.76} & \textbf{6.33} \\
		\hline
	\end{tabular}
	\vspace{0mm}
\end{table*}

\begin{table*}[!htb]
	\centering
	
	\caption{ROUGE evaluation on dataset Yelp-2016.}
	\label{tab:rouge-yelp}
	\begin{tabular}{|c|c|c|c|c|c|c|c|c|c|c|c|c|}
		\hline
		\multirow{2}{*}{\textbf{Methods}} &
		\multicolumn{3}{c|}{\textbf{ROUGE-1}} &
		\multicolumn{3}{c|}{\textbf{ROUGE-2}} &
		\multicolumn{3}{c|}{\textbf{ROUGE-L}} &
		\multicolumn{3}{c|}{\textbf{ROUGE-SU4}} \\
		\cline{2-13}
		& R & P & F1 & R & P & F1& R & P & F1& R & P & F1 \\
		%\hline
		%LexRank-T & 20.26 & 21.00 & 19.56
		%& 10.73 & 11.17 & 10.68
		%& 18.72 & 19.20 & 18.03
		%& 12.49 & 12.94 & 12.12 \\
		\hline
		LexRank & 11.32 & 11.16 & 11.04
		& 1.32 & 1.34 & 1.31
		& 10.33 & 10.16 & 10.06
		& 3.41 & 3.38 & 3.26 \\
		%\hline
		%RMR-T & 15.37 & 13.79 & 13.87
		%& 5.19 & 4.99 & 5.02
		%& 13.85 & 14.87 & 12.54
		%& 7.29 & 6.65 & 6.62 \\
		\hline
		RMR$_t$ & 11.17 & 10.25 & 10.54
		& 2.25 & 2.16 & 2.19
		& 10.22 & 9.39 & 9.65
		& 3.88 & 3.66 & 3.72 \\
		%\hline
		%CTR-T & 15.80 & 14.19 & 14.26
		%& 5.82 & 5.57 & 5.61
		%& 14.29 & 14.87 & 12.92
		%& 7.78 & 7.14 & 7.11 \\
		\hline
		CTR$_t$ & 10.74 & 9.95 & 10.19
		& 2.21 & 2.14 & 2.15
		& 9.91 & 9.19 & 9.41
		& 3.96 & 3.64 & 3.70 \\
		\hline
		\textbf{NRT} & 9.39 & \textbf{17.75} & \textbf{11.64}
		& 1.83 & \textbf{3.39} & \textbf{2.22}
		& 8.70 & \textbf{16.27} & \textbf{10.74}
		& 3.01 & \textbf{7.06} & \textbf{3.78} \\
		\hline
	\end{tabular}
\end{table*}

For most of the datasets, our NRT model does not outperform
the baselines on Recall.
There are several reasons: (1) The ground truth tips used in the training set are very short, only about 10-word length on average.
Naturally, the model trained using this dataset cannot generate long sentence.
(2) The mechanism of typical beam search algorithm makes the model favor short sentences.
(3) The comparison models are extraction-based approaches and these models favor to extract long sentence,
although we add a length (i.e., 20 words) restriction on them.
%For example, LexRank employs Pagerank algorithm to estimate the importance of sentences from a sentence graph. And sentence similarity is used to initialize the weight of edges. Intuitively,  sentence with more words will connect with more other sentences, which will get a higher salience score. We also utilize similarity calculation to extract tips for CTR$_t$ and RMR$_t$, so they performs similarly with LexRank on preferring long sentences.

\begin{figure}[!t]
	\centering
	\begin{subfigure}[t]{0.5\columnwidth}
		\centering
		\includegraphics[height=1.4in]{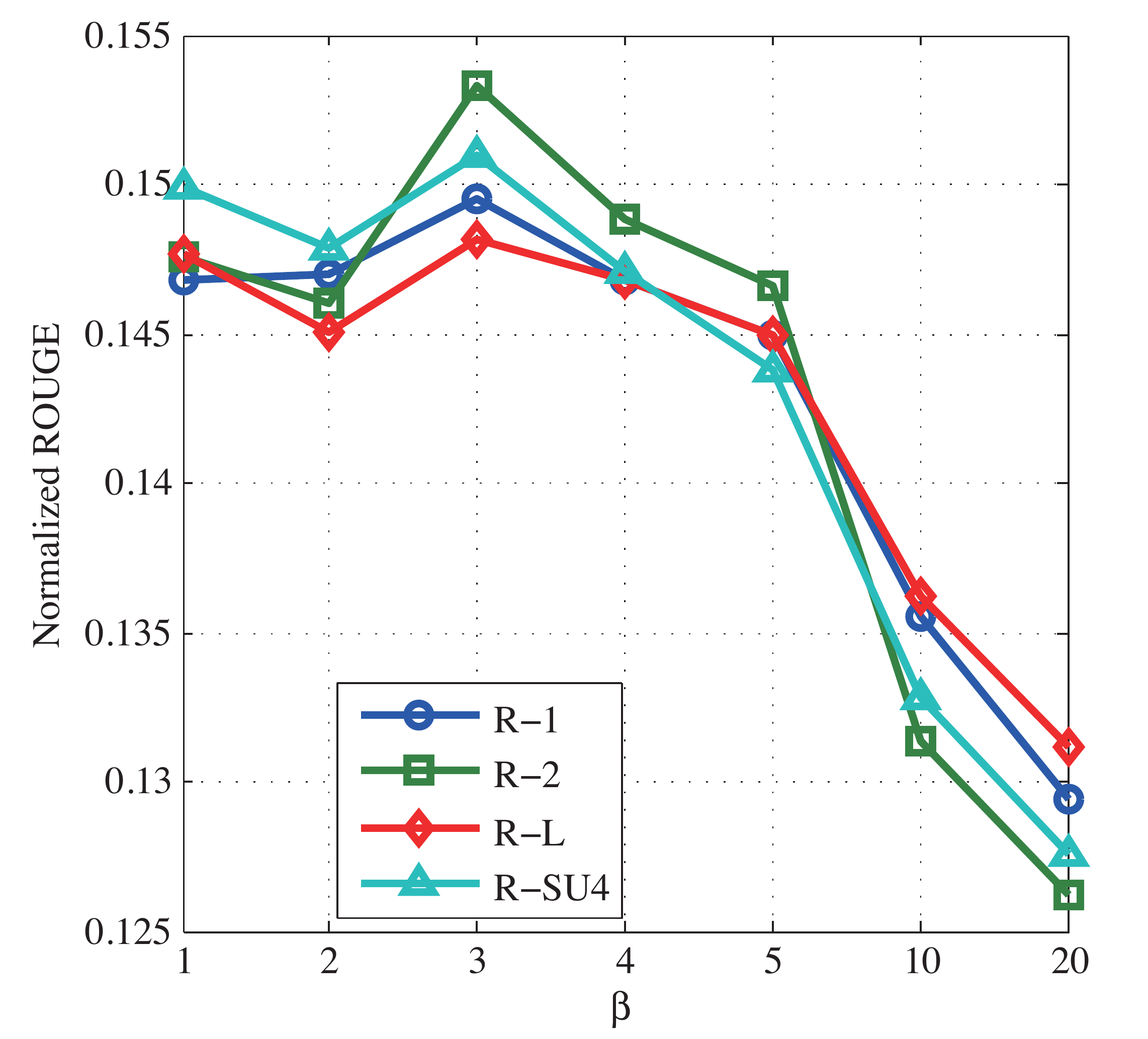}
		\caption{Electronics.}
	\end{subfigure}%
	~
	\begin{subfigure}[t]{0.5\columnwidth}
		\centering
		\includegraphics[height=1.4in]{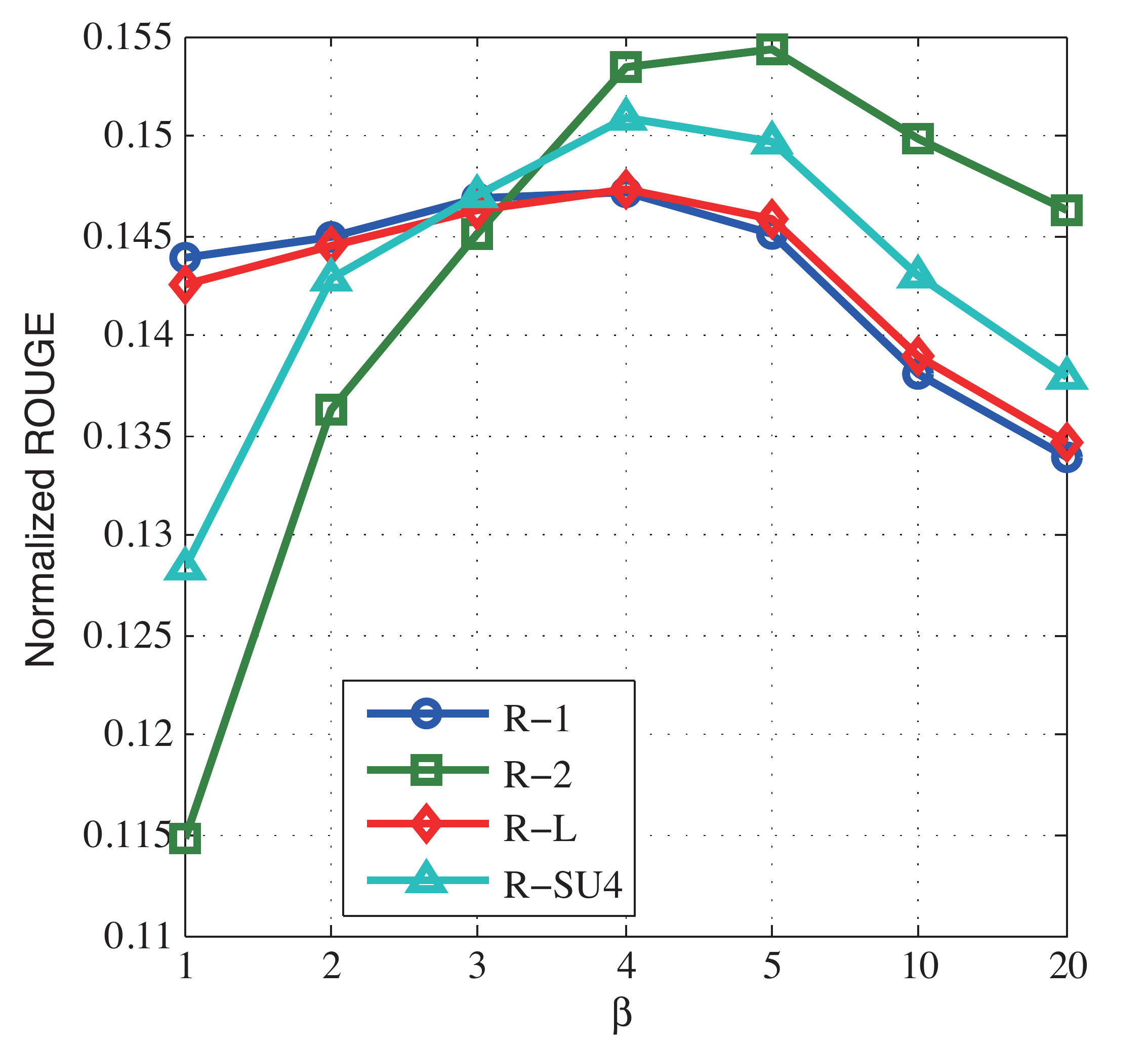}
		\caption{Movies\&TV. }
	\end{subfigure}
	\caption{Effectiveness of beam size $\beta$ on the validation set.}
	\label{fig:tune_beam}
	\vspace{0mm}
\end{figure}

We investigate the performance of different beam size $\beta$ used in the beam search algorithm.
The relationship between ROUGE and $\beta$ on two validation sets of Electronics and Movies\&TV is shown in Figure~\ref{fig:tune_beam}.
We test $\beta \sim \{1\sim5, 10, 20\}$ and find that when $\beta = 3\sim5$ our model can achieve the best performance of tips generation.

Inspired by \cite{wu2016google}, we make use of Length-Normalization (LN) to adjust the  log-probability in the beam search algorithm to make the beam search algorithm also consider long sentences:
\begin{equation}
LN(s) = \frac{{{{(n + |s|)}^\alpha }}}{{{{(n + 1)}^\alpha }}}
\end{equation}
where $s$ is the decoded sequence, $n = 2$, and $\alpha = 0.6$.
We conduct several experiments to verify the effectiveness of LN.
The comparison results are shown in Table~\ref{tbl:rouge-ln}, where F1-measures of ROUGE evaluation metrics are reported.
It is obvious that our model NRT with LN is much better than the one without LN.

\begin{table}[!t]
	\centering
	\caption{Effectiveness of Length-Normalization (LN). R-* refers to ROUGE-*.}
	\label{tbl:rouge-ln}
	\begin{tabular}{l c c c c c}
		\hline
		\textbf{Dataset} & \textbf{Method} & \textbf{R-1} & \textbf{R-2} & \textbf{R-L} & \textbf{R-SU4} \\
		\hline
		Electronics	& NRT w/o LN       & 13.36 & 2.65 & 12.34 & 4.56 \\
		& \textbf{NRT}     & \textbf{13.72} & \textbf{2.68} & \textbf{12.57} & \textbf{4.66} \\
		\hline
		Movies\&TV	& NRT w/o LN        & 14.86 & 3.72 & 13.76 & 5.46 \\
		& \textbf{NRT}     & \textbf{15.21} & \textbf{4.00} & \textbf{13.90} & \textbf{5.71} \\
		\hline
	\end{tabular}
	\vspace{0mm}
\end{table}

\subsection{Case Analysis (RQ3)}
\label{sec:exp:case}

For the purpose of analyzing the linguistic quality and the sentiment correlation between the predicted ratings and the generated tips, we selected some real cases form different domains. The results are listed in Table~\ref{tbl:case}.
Although our model generates tips in an abstractive way, tips' linguistic quality is quite good.

For the sentiment correlation analysis, we also choose some generated tips with negative sentiment. Take the tips ``Not as good as i expected.'' as an example, our model predicts a rating of $2.25$, which clearly shows the consistent sentiment. The ground truth tips of this example is ``Jack of all trades master of none. '', which also conveys a negative sentiment. One interesting observation is that its ground truth rating is the full mark $5$, which we guess, may be clicked by a fat finger.
Nevertheless, our model can generate a consistent sentiment between this case's rating and tips. Another generated tips ``What a waste of time and money.'' with a negative predicted rating of $1.46$ also demonstrates this property.

There are also some bad cases. For example, the predicted rating of the generated tips ``Not bad for the price.'' is $4.34$, which is a positive polarity. But the sentiment of the generated tips is neutral, consistent with the ground truth.
Generally speaking, our model can achieve satisfactory performance on both rating prediction and abstractive tips generation.

\begin{table}[!t]
	\centering
	\caption{Examples of the predicted ratings and the generated tips. The first line of each group shows the generated rating and tips.
		The second line shows the ground truth.}%
	\label{tbl:case}%
	\begin{tabular}{ c | p{6.4cm}}
		\hline
		\textbf{Rating} & \ \ \ \ \ \  \  \ \ \ \ \ \ \ \ \ \ \ \ \ \ \ \ \ \ \ \ \ \ \textbf{Tips} \\
		\hline
		\textbf{\textit{4.64}} & \textbf{\textit{This is a great product for a great price.}} \\
		5 & Great product at a great price.  \\
		
		\hline
		\textbf{\textit{4.87}} &\textbf{\textit{ I purchased this as a replacement and it is a perfect fit and the sound is excellent.}} \\
		5 & Amazing sound.  \\
		
		\hline
		\textbf{\textit{4.69}} &\textbf{\textit{ I have been using these for a couple of months.}} \\
		4 & Plenty of wire gets signals and power to my amp just fine quality wise.  \\
		
		\hline
		\textbf{\textit{4.87}} &\textbf{\textit{ One of my favorite movies.}} \\
		5 & This is a movie that is not to be missed.  \\

		\hline
		\textbf{\textit{4.07}} &\textbf{\textit{ Why do people hate this film.}} \\ 4 & Universal why didnt your company release this edition in 1999. \\
		
		\hline
		\textbf{\textit{2.25}} &\textbf{\textit{ Not as good as i expected.}} \\ 5 & Jack of all trades master of none. \\
		
		\hline
		\textbf{\textit{1.46}} &\textbf{\textit{ What a waste of time and money.}} \\ 1 & The coen brothers are two sick bastards.  \\
		\hline
		\hline
		
		\textbf{\textit{4.34}} & \textbf{\textit{Not bad for the price.}} \\
		3 & Ended up altering it to get rid of ripples. \\

		\hline
	\end{tabular}
	\vspace{-3mm}
\end{table}

\begin{comment}
\hline
\textbf{\textit{4.60}} &\textbf{\textit{I am not a big fan of the Coen brothers but this movie was great.}} \\
5 & Gritty rough powerhouse of a drama well made and beautifully acted. \\
\end{comment}

%\section{Conclusions}
\section{Conclusions}

We propose a deep learning based framework named \textbf{NRT} which can simultaneously predict precise ratings and generate abstractive tips with good linguistic quality simulating user experience and feelings.
For abstractive tips generation, GRU with context information is employed to ``translate'' user and item latent factors into a concise sentence.
All the neural parameters as well as the latent factors for users and items are learnt by a multi-task learning approach in an end-to-end training paradigm.
Experimental results on benchmark datasets show that NRT achieves better performance than the state-of-the-art models on both tasks of rating prediction and abstractive tips generation.
The generated tips can vividly predict the user experience and feelings.

%\balance
\bibliographystyle{ACM-Reference-Format}
\bibliography{sigproc}

\end{document}